\documentclass{article}
\usepackage{smoothie-qwen}

\PassOptionsToPackage{table}{xcolor}
\usepackage{etoolbox}
\usepackage[utf8]{inputenc}

\newtoggle{arxiv}
\toggletrue{arxiv}

\usepackage{booktabs}
\usepackage{graphicx}
\usepackage{enumitem}
\usepackage{wrapfig}
\usepackage{algorithm}
\usepackage{algpseudocode}

\usepackage{microtype}
\usepackage{amsmath}
\usepackage{colortbl}
\definecolor{lightgray}{rgb}{0.9,0.9,0.9}
\usepackage{caption}
\usepackage{subcaption}
\usepackage{xcolor}
\usepackage{setspace}
\usepackage{url}
\usepackage{multirow}
\usepackage{colortbl}
\usepackage{tabularx}
\usepackage{blindtext}
\usepackage{pgfplots}
\pgfplotsset{compat=1.18} 
\usepackage{tikz}
\usetikzlibrary{er,positioning,bayesnet}
\usepackage{makecell}
\usepackage{tipa}
\usepackage{siunitx}
\usepackage{nicefrac}
\usepackage{tocloft}
\usepackage{listings}
\usepackage[raster,skins]{tcolorbox} 
\usepackage{xltabular}
\usepackage{adjustbox}
\usepackage{xurl}
\usepackage{kotex}
\usepackage{natbib}
\usepackage{threeparttable}
\usepackage{hyperref}
\usepackage{algorithm}
\usepackage{algpseudocode}

\title{
    \includegraphics[width=0.3\textwidth]{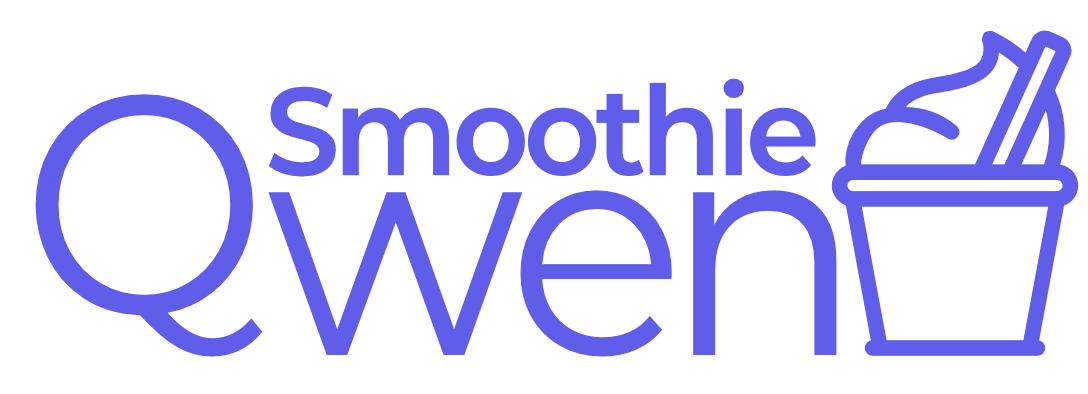} \\ \vspace{0.5cm}
    Smoothie-Qwen: Post-Hoc Smoothing to Reduce Language Bias in Multilingual LLMs
}
\author{SeungWon Ji, Jungyup Lee, Jemin Kim, Sang Park, SeungJae Lee}
\date{\today}

\begin{document}

\maketitle

\begin{abstract}
Multilingual large language models (LLMs) often exhibit language confusion, a tendency to generate responses in a dominant language irrespective of the prompt's language. To address this, we propose \textbf{Smoothie-Qwen}, a lightweight, post-hoc method that mitigates language bias without retraining. This technique selectively adjusts token-level output probabilities to effectively suppress undesired language generation. Applied to the Qwen model, our method reduces unintended Chinese output by \textbf{over 95\%} while preserving task accuracy on multilingual benchmarks. This work provides a practical and efficient solution for enhancing the language controllability of LLMs, making them more reliable for global applications.
\end{abstract}
\section{Introduction}

In recent years, the field of artificial intelligence has made significant strides with the emergence of powerful multilingual large language models (LLMs) such as Qwen\citep{yang2025qwen3} and LLaMA\citep{grattafiori2024llama}. Trained on massive datasets spanning dozens of languages, these models have enabled a wide range of multilingual applications. However, they have also revealed a critical limitation known as language confusion—a phenomenon in which the model tends to generate responses in a dominant language (e.g., Chinese), even when prompted in another language\citep{marchisio2024understanding}. This issue arises from imbalances in the training data and is not a bug, but rather a predictable outcome of standard training paradigms that learn and reinforce the distributional biases present in the data. This language confusion is a critical barrier to global LLM adoption because it undermines reliability and user trust through unpredictable outputs.

To address this problem, we propose \textbf{Smoothie-Qwen}, a lightweight, post-hoc adjustment method that can be applied without retraining. Unlike rigid approaches that block the generation of specific tokens entirely, Smoothie-Qwen adopts a \textbf{smoothing-based strategy} that softly modulates token probability distributions at the model’s final output layer (\texttt{lm\_head}). This approach allows for a nuanced adjustment of the trade-off between suppressing undesired tokens and preserving the model's overall expressive capacity, as token weights are downscaled but never entirely eliminated. As a result, users can efficiently mitigate undesired behavioral biases and guide the model toward producing more natural and balanced multilingual outputs—with minimal computational cost and no need to alter the original training pipeline.
\section{Related Work}

Research has documented extensive biases in large language models (LLMs), from social stereotypes like gender and nationality bias~\citep{zhou2024evaluating,kamruzzaman2024banstereoset} to functional issues such as \textbf{“language confusion,”} where models default to a dominant language irrespective of the user's prompt~\citep{marchisio2024understanding}. Mitigation strategies are broadly categorized as pre-processing, in-training, and post-processing~\citep{gallegos2024bias}. While effective, pre-processing and in-training solutions like data augmentation or supervised fine-tuning (SFT) are often impractical for large-scale, deployed models due to the prohibitive cost and data access limitations of retraining.

This has led to a growing focus on post-hoc interventions that require no retraining. These range from flexible but sometimes inconsistent approaches like prompt engineering and zero-shot self-debiasing~\citep{gallegos2024self}, to output correction frameworks~\citep{zhong2024harnessing} and direct \textbf{model editing} techniques that surgically modify model parameters~\citep{pan2023automatically}. Our work, \textbf{Smoothie-Qwen}, contributes to the post-hoc model editing literature by proposing a novel technique. Instead of altering factual knowledge, it introduces a lightweight, probabilistic smoothing of the final output layer, offering a unique balance of efficiency and fine-grained control to specifically address language bias.
\section{The Smoothie-Qwen Method: A Probabilistic Smoothing Framework}

\subsection{Conceptual Foundation: Post-Hoc Bias Mitigation via Direct Weight Modification}

\textbf{Smoothie-Qwen} is a post-processing technique designed to mitigate language bias by directly modifying a fully trained model. Unlike pre-processing methods that alter datasets or in-training approaches that demand costly retraining, Smoothie-Qwen operates directly on deployed models. Its core advantage lies in its efficiency and accessibility, as it can be applied to any publicly available model without requiring access to original training data or performing additional training.

The key idea is to selectively reduce the generation probability of certain language tokens by directly editing the weights in the model’s final output layer, \texttt{lm\_head}, which determines the token-level probability distribution. This constitutes a form of \textbf{post-hoc model editing}, allowing for fine-grained behavioral control over a pretrained language model without compromising its general capabilities.

\begin{figure}[h!]
    \centering
    \includegraphics[width=0.85\textwidth]{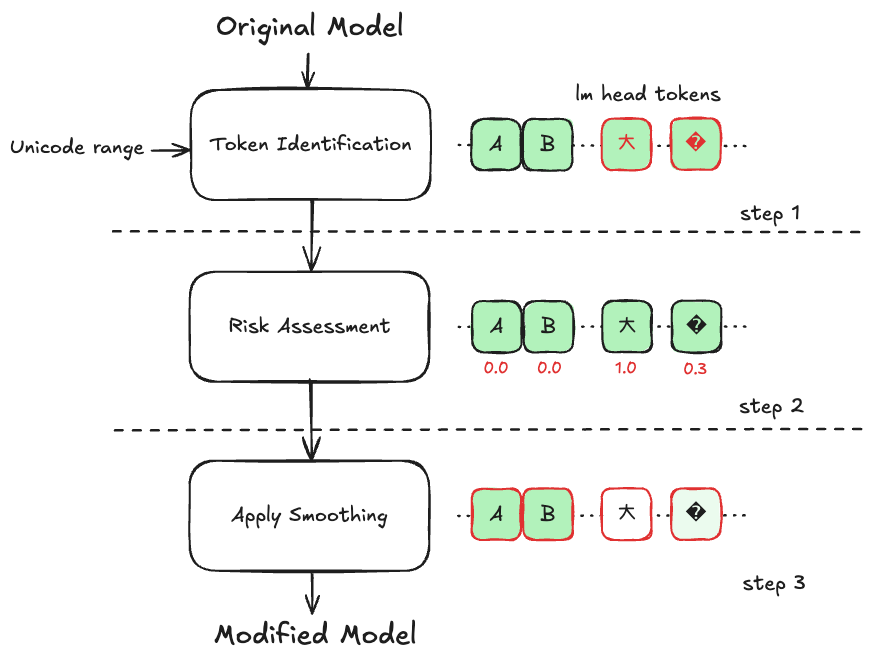}
    \caption{
        Overview of the Smoothie-Qwen workflow. The process begins by identifying target tokens through Unicode range and subword pattern analysis. Each token is then assigned a probabilistic risk score, which guides the final step: applying a targeted smoothing function to downscale the weights of high-risk tokens in the final \texttt{lm\_head} layer. This figure illustrates how the original model is transformed into a debiased version through these lightweight post-hoc modifications.
    }
    \label{fig:flow_chart}
\end{figure}

\subsection{Step 1: Identifying Target Tokens via Unicode and Subword Analysis}

The first step in the Smoothie-Qwen algorithm is to comprehensively identify tokens corresponding to a specific target language (e.g., Chinese) slated for suppression. This process consists of two main procedures:

\begin{itemize}
    \item \textbf{Target Token Identification Based on Unicode Ranges}
    The most direct approach involves scanning the model’s entire vocabulary using user-defined Unicode ranges associated with the target language. For instance, to suppress Chinese tokens, a user might provide the range U+4E00 to U+9FFF. The algorithm iterates through the vocabulary, and any token falling within this range is flagged as a high-risk candidate for the subsequent smoothing stage.

    \item \textbf{Detection of Broken Tokens}
    Modern Byte-Pair Encoding (BPE) tokenizers can split a single character into multiple subword fragments. Consequently, several “broken” tokens may combine to form target-language characters. For example, the replacement character  (U+FFFD) often falls into this category. To ensure comprehensive coverage, Smoothie-Qwen explicitly identifies these potentially problematic fragments and includes them in the risk assessment process.
\end{itemize}

In practice, the Qwen3 tokenizer contains 151,643 vocabulary entries. Among these, 26,153 tokens (17.25\%) fall within the defined Chinese Unicode range, and 1,457 tokens (0.96\%) are identified as broken tokens. By addressing both direct and fragmented tokens, Smoothie-Qwen establishes a comprehensive foundation for its risk analysis.

\subsection{Step 2: Probabilistic Risk Assessment for Each Token}

The second step quantifies the probabilistic risk of each token by assigning it a \texttt{risk\_score} between 0.0 and 1.0, representing its likelihood of contributing to language bias. The score is determined by the token's classification from Step 1:

\begin{itemize}
    \item \textbf{Target tokens}: Tokens directly identified as part of the target language (e.g., Chinese characters) are considered high-risk and are assigned a \texttt{risk\_score} of 1.0.

    \item \textbf{Broken tokens}: The risk for these tokens is estimated based on their potential to form target-language characters when combined with other fragments. This potential is evaluated using an N-gram sampling approach, where the N-gram size (e.g., 2, 3, or 4) is a configurable parameter.

    \item \textbf{Safe tokens}: Tokens unrelated to the target language (e.g., English tokens) are considered safe and receive a \texttt{risk\_score} of 0.0.

\end{itemize}

For instance, to evaluate a broken token \texttt{A} with a 2-gram analysis, the algorithm might pair it with 100 randomly selected tokens to form 2-gram combinations (e.g., A+B, A+C). If 12 of these combinations result in valid characters from the target language, the \texttt{risk\_score} of \texttt{A} is set to 0.12.

This sampling strategy avoids the computational burden of exhaustively checking all possible combinations (e.g., $1{,}457^n$), providing an efficient yet effective approximation of token risk while keeping the process tractable.

This process concludes by assigning a probabilistic \texttt{risk\_score} to every token in the vocabulary. This score serves as the key signal for determining the intensity of weight reduction in the subsequent smoothing stage.

\subsection{Step 3: Smoothing Transformation}

The third and final step uses the \texttt{risk\_score} from Step 2 to guide a smoothing transformation that directly modifies the weights in the \texttt{lm\_head} layer. The goal is to reduce the generation probability of high-risk tokens, thereby mitigating language bias.

\subsubsection{Characteristics and Considerations of Weight Scaling}

A foundational concept in this step is weight scaling: reducing a token's generation probability by multiplying its corresponding \texttt{lm\_head} weights by a factor $S$, where $0 < S \leq 1$. This directly alters the model’s logits—the raw outputs before softmax normalization—but its effect depends on the logit's original sign:

\begin{itemize}
    \item \textbf{Positive Logits (e.g., +10):} Scaling reduces the value (e.g., to +5), directly lowering the token’s generation probability as intended.
    \item \textbf{Negative Logits (e.g., -10):} Scaling reduces the magnitude (e.g., to -5), moving the value closer to zero. This can paradoxically increase the token’s relative probability after softmax normalization.
\end{itemize}

This dynamic introduces a potential trade-off: while high-risk tokens with positive logits are suppressed correctly, those with negative logits could experience a slight probability increase.

Fortunately, this trade-off is largely theoretical in our context. Empirical analysis confirms that high-risk tokens—such as those from the target language—overwhelmingly have positive logits. This ensures that the smoothing transformation is effective at reducing their generation probability, while any unintended side effects on other tokens are minimal and manageable.

\subsubsection{Risk-Aware Scaling via Non-Linear Smoothing}

While the basic principle of weight scaling is effective, it lacks fine-grained control. Applying a single, uniform scaling factor to all target tokens fails to account for their varying degrees of risk, potentially under-suppressing highly problematic tokens while over-penalizing less risky ones. To address this and achieve more precise control, Smoothie-Qwen employs a non-linear smoothing function that dynamically adjusts the scaling factor $S$ based on each token's individual \texttt{risk\_score}.

The function is defined as follows:
\begin{equation} \label{eq:smoothing}
S = 1 - (1 - \textit{min\_scale}) \times \frac{\log\left(1 + (\textit{smoothness} - 1) \times \texttt{risk\_score}\right)}{\log(\textit{smoothness})}
\end{equation}

This logarithmic function ensures that tokens with a higher \texttt{risk\_score} are assigned a smaller scaling factor—closer to \texttt{min\_scale}—leading to stronger suppression. Conversely, tokens with a low \texttt{risk\_score} are scaled by a factor close to 1.0, thereby minimally altering their original weight.

The two key hyperparameters in this formulation are:

\begin{description}
    \item[\textbf{min\_scale}] 
    This parameter sets the minimum possible value for the scaling factor $S$. It guarantees that even the highest-risk tokens retain a fraction of their original weight, preventing it from being zeroed out completely. This safeguard is especially important in scenarios like translation, where a suppressed language may still be required in the output.

    \item[\textbf{smoothness}] 
    This parameter controls the steepness of the suppression curve. Higher values result in more aggressive downscaling across a broader range of \texttt{risk\_score}s, but overly aggressive settings risk penalizing valid language generation.
\end{description}

Together, these hyperparameters provide a flexible toolkit for navigating the trade-off between bias mitigation and model performance, allowing users to tune the intensity and floor of the suppression to fit their specific needs.

\begin{figure}[h!]
    \centering
    \includegraphics[width=0.8\textwidth]{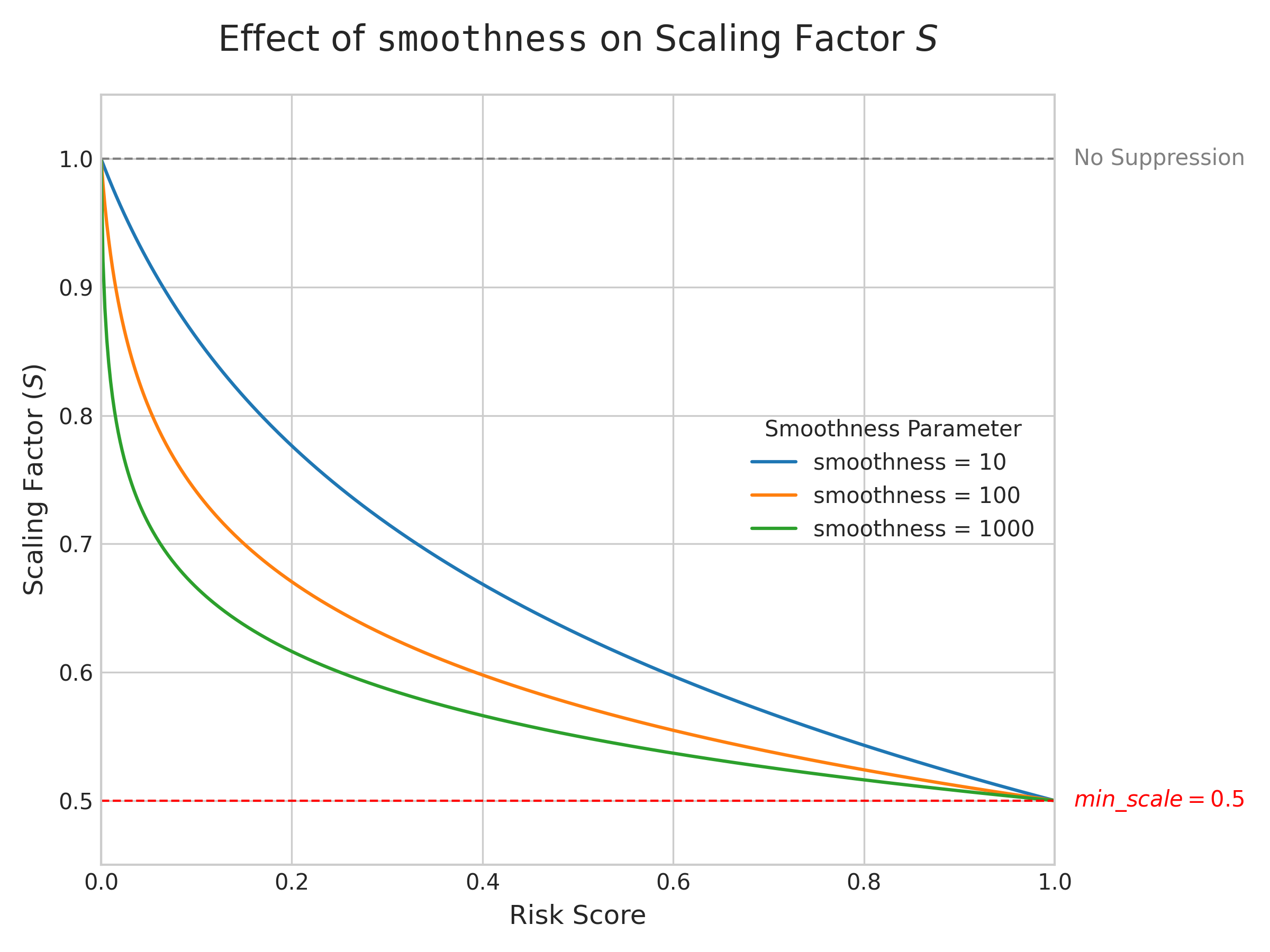}
    \caption{
        Visualization of the non-linear smoothing function, illustrating the relationship between the \textit{risk\_score} (x-axis) and the final scaling factor $S$ (y-axis) for a fixed \textit{min\_scale} of 0.5. The plot compares three different \textit{smoothness} values (10, 100, and 1000). As shown, all curves start at $S=1$ (no suppression) when the \textit{risk\_score} is 0 and converge toward the \textit{min\_scale} of 0.5 as the \textit{risk\_score} approaches 1. A higher \textit{smoothness} value results in a steeper, more aggressive curve, applying significant suppression even to tokens with moderate risk scores. This demonstrates how the \textit{smoothness} parameter provides fine-grained control over the intensity of the debiasing transformation.
    }
    \label{fig:smoothing_plot}
\end{figure}

\subsection{Implementation and Application}

Smoothie-Qwen is a lightweight, training-free method that operates post-hoc, requiring no architectural changes. It functions by directly adjusting the model’s final projection layer (\texttt{lm\_head}) using token-level risk scores and smoothing parameters. The entire process is formally defined in Algorithm~\ref{alg:smoothie_qwen}.

\begin{algorithm}
\caption{Smoothie-Qwen: Risk-Aware Suppression of Target Language Tokens}
\label{alg:smoothie_qwen}
\begin{algorithmic}[1]
\Require Pretrained language model $M$, vocabulary $\mathcal{V}$, Unicode range $\mathcal{U}$, \textit{min\_scale}, \textit{smoothness}
\Ensure Modified model $M'$ with smoothed \texttt{lm\_head} weights

\State Identify target tokens $\mathcal{T}_{\text{target}}$ and broken tokens $\mathcal{T}_{\text{broken}}$ from $\mathcal{V}$
\State Initialize an empty dictionary $\texttt{risk\_score}$

\ForAll{token $t$ in $\mathcal{V}$} \Comment{First, determine the risk score for all tokens}
    \If{$t \in \mathcal{T}_{\text{target}}$}
        \State $\texttt{risk\_score}[t] \gets 1.0$ \Comment{Target tokens}
    \ElsIf{$t \in \mathcal{T}_{\text{broken}}$}
        \State Estimate $\texttt{risk\_score}[t]$ via $n$-gram sampling \Comment{Broken tokens}
    \Else
        \State $\texttt{risk\_score}[t] \gets 0.0$ \Comment{Safe tokens}
    \EndIf
\EndFor

\ForAll{token $t$ in $\mathcal{V}$} \Comment{Then, apply smoothing based on the finalized risk scores}
    \State $r \gets \texttt{risk\_score}[t]$
    \If{$r > 0$}
        \State $S \gets 1 - (1 - \textit{min\_scale}) \cdot \dfrac{\log(1 + (\textit{smoothness} - 1) \cdot r)}{\log(\textit{smoothness})}$
        \State Get weight vector $w_t$ for token $t$ from $M.\texttt{lm\_head}$
        \State Update weight vector in $M'.\texttt{lm\_head}$: $w'_t \gets S \cdot w_t$
    \EndIf
\EndFor

\State \Return Modified model $M'$
\end{algorithmic}
\end{algorithm}

This concise procedure highlights the simplicity and modularity of the Smoothie-Qwen framework. As it operates entirely at the post-processing stage, it can be seamlessly applied to any compatible pretrained model without modifying upstream training pipelines. To promote reproducibility and broader adoption, all source code for Smoothie-Qwen is publicly available on GitHub\footnote{\url{https://github.com/dnotitia/smoothie-qwen}}. In addition, we provide ready-to-use models on Hugging Face. These models are based on the official Qwen3 models, with our smoothing transformation pre-applied\footnote{\url{https://huggingface.co/collections/dnotitia/smoothie-qwen3-6811896ebb3a255de7b5b437}}.

\begin{figure}[h!]
    \centering
    \includegraphics[width=\textwidth]{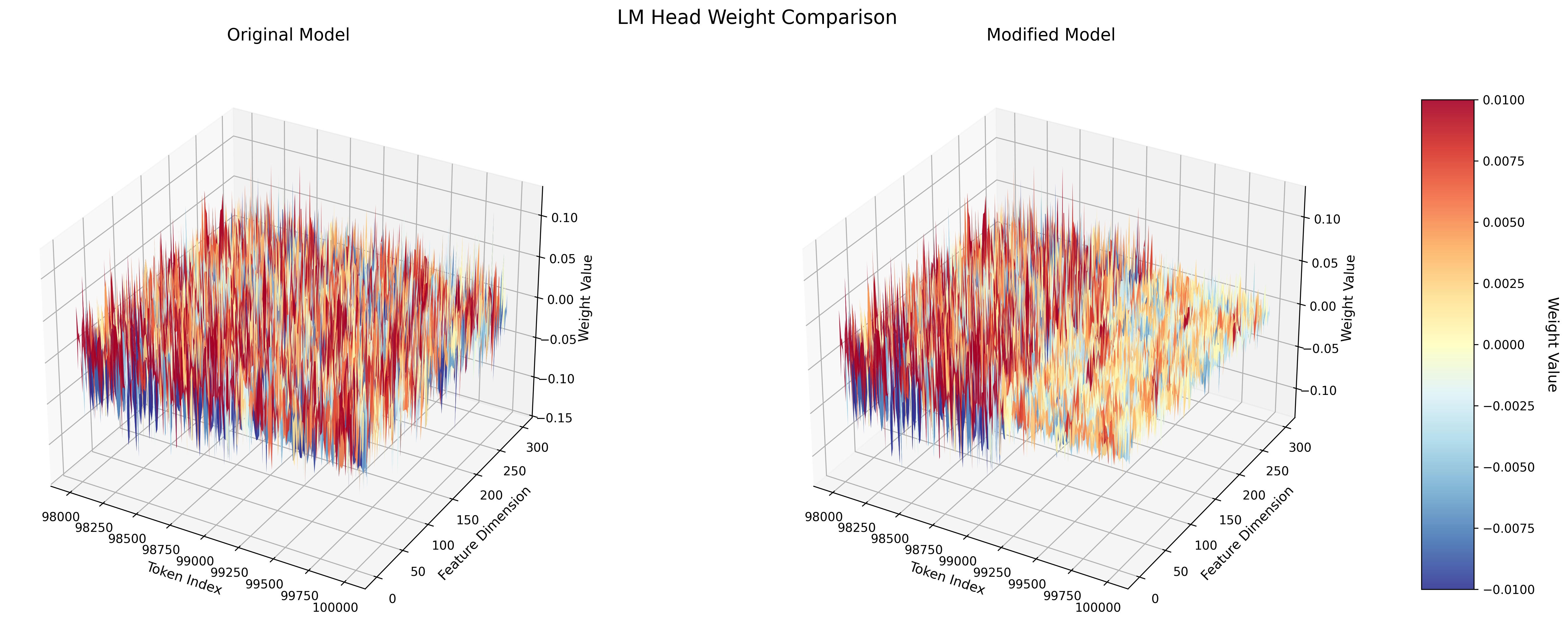}
    \caption{Visual comparison of \texttt{lm\_head} weights before (left) and after (right) the smoothing transformation. This figure shows a specific slice of token indices (98,000-100,000) that contains a mix of Chinese and non-Chinese tokens. In the "Modified Model" plot, the weights corresponding to Chinese tokens (which have a high \textit{risk\_score}) are scaled towards zero. This causes both the originally high (hot colors) and low (cool colors) weight values to converge toward the center, resulting in a distinctly \textbf{``flattened''} surface. In contrast, the weights of adjacent, non-Chinese tokens remain largely unchanged, clearly illustrating the targeted nature of the smoothing process.}
    \label{fig:weight_comparison}
\end{figure}
\section{Experiments}

\subsection{Experimental Setup}

\subsubsection{Base Model and Suppression Target}
Our experiments are conducted on the \textbf{Qwen2.5-Coder-14B-Instruct} model\citep{hui2024qwen2}, with the primary objective of suppressing unwanted \textbf{Chinese language generation bias}.

Upon analyzing the model's full vocabulary of 151,643 tokens, we identified:
\begin{itemize}
    \item \textbf{26,153 target Chinese tokens} (17.25\%) that fall within the Chinese Unicode range.
    \item \textbf{1,457 broken tokens} (0.96\%) that may contribute to Chinese token formation via subword combinations.
\end{itemize}
These proportions highlight a non-trivial presence of target-language tokens, establishing a clear motivation for bias mitigation.

\subsubsection{Evaluation Datasets and Metrics}
To assess both \textbf{bias mitigation effectiveness} and the \textbf{preservation of core model functionality}, we utilized EleutherAI’s \texttt{lm-evaluation-harness}~\citep{eval-harness} with the following datasets and metrics:

\begin{itemize}
    \item \textbf{Custom Chinese Elicitation Dataset:} 
    To measure unintended Chinese generation in a controlled environment, we constructed a synthetic dataset of 1,000 prompt-response pairs. 
    Each prompt explicitly asks for a Chinese translation of a Korean word (e.g., \textit{"중국어로 '냉장고'는 무엇인가요?"}). 
    To ensure domain diversity, prompts were drawn from 26 categories—including numbers, food, sports, and family—and were specifically designed to reliably elicit Chinese output.
    \begin{itemize}
        \item \textbf{chin\_prom:} 
        Measures how effectively Chinese responses are suppressed in this elicitation setting.
    \end{itemize}

    \item \textbf{KMMLU (Korean-MMLU):} 
    A Korean-language benchmark derived from MMLU\citep{son2024kmmlu}, covering academic domains such as Computer Science (CS) and Industrial Engineering (IE). For each domain, we applied two complementary evaluation modes:
    \begin{itemize}
        \item \textbf{chin\_cs / chin\_ie:} 
        Using a generative evaluation setup, these metrics measure the proportion of Chinese content in free-form Korean responses, reflecting suppression effectiveness.

        \item \textbf{acc\_cs / acc\_ie:} 
        Following the original multiple-choice protocol, these metrics assess whether the model’s task-solving accuracy is preserved after applying Smoothie-Qwen.
    \end{itemize}
\end{itemize}

For all \texttt{chin\_*} metrics, a higher score (closer to 1.0) indicates more effective suppression of unintended Chinese generation.

\subsection{Quantitative Evaluation}

This section quantitatively evaluates the impact of the key hyperparameters, \texttt{min\_scale} and \texttt{smoothness}, on both Chinese suppression rates and Korean task accuracy.

\subsubsection{Effect of \texttt{min\_scale}}

The \texttt{min\_scale} parameter sets the floor for weight suppression, with lower values corresponding to more aggressive reduction. For this experiment, we fixed \texttt{smoothness} at 10.0 while varying \texttt{min\_scale} from 1.0 (baseline) down to 0.1. The results are presented in Table~\ref{tab:min_scale_results}.

\begin{table}[h!]
\centering
\caption{Effect of \texttt{min\_scale} on Chinese suppression and task accuracy}
\label{tab:min_scale_results}
\begin{tabular}{cccccc}
\hline
\textbf{min\_scale} & \textbf{chin\_prom} & \textbf{chin\_cs} & \textbf{chin\_ie} & \textbf{acc\_cs} & \textbf{acc\_ie} \\
\hline
1.0 (base) & 0.190 & 0.995 & 0.990 & 0.715 & 0.385 \\
0.9 & 0.250 & 0.995 & 1.000 & 0.710 & 0.395 \\
0.8 & 0.375 & 0.995 & 1.000 & 0.710 & 0.395 \\
0.7 & 0.605 & 0.995 & 1.000 & 0.710 & 0.395 \\
0.6 & 0.875 & 0.995 & 1.000 & 0.710 & 0.395 \\
\textbf{0.5} & \textbf{0.950} & \textbf{0.995} & \textbf{1.000} & \textbf{0.710} & \textbf{0.395} \\
0.4 & 0.965 & 0.995 & 1.000 & 0.710 & 0.395 \\
0.3 & 0.980 & 0.995 & 1.000 & 0.710 & 0.395 \\
0.2 & 0.985 & 1.000 & 1.000 & 0.710 & 0.395 \\
0.1 & 0.990 & 1.000 & 1.000 & 0.710 & 0.395 \\
\hline
\end{tabular}
\end{table}

As presented in the table, lowering \texttt{min\_scale} has a direct and significant impact on the suppression rate for the Chinese elicitation dataset (\texttt{chin\_prom}), which increases from 0.190 to 0.990. Notably, at \texttt{min\_scale} = 0.5, a high suppression rate of 95\% is achieved with almost no measurable degradation in task performance on the KMMLU benchmarks (\texttt{acc\_cs}, \texttt{acc\_ie}). This finding demonstrates that Smoothie-Qwen can effectively mitigate language bias without compromising the model’s core reasoning abilities. Although suppression rates continue to improve as \texttt{min\_scale} approaches 0.1, the gains become marginal. Therefore, we identify \texttt{min\_scale} = 0.5 as a \textbf{highly efficient and sufficient choice} for practical applications.

\subsubsection{Effect of \texttt{smoothness}}

The \texttt{smoothness} hyperparameter controls the steepness of the suppression curve, thereby governing the aggressiveness of the transformation. For this experiment, we fixed \texttt{min\_scale} at 0.5 and varied \texttt{smoothness} from 1.1 to 1000.0. The results are presented in Table~\ref{tab:smoothness_results}.

\begin{table}[h!]
\centering
\caption{Effect of \texttt{smoothness} on Chinese suppression and task accuracy}
\label{tab:smoothness_results}
\begin{tabular}{cccccc}
\hline
\textbf{smoothness} & \textbf{chin\_prom} & \textbf{chin\_cs} & \textbf{chin\_ie} & \textbf{acc\_cs} & \textbf{acc\_ie} \\
\hline
1.1 & 0.920 & 0.995 & 1.000 & 0.710 & 0.395 \\
\textbf{10.0} & \textbf{0.950} & \textbf{0.995} & \textbf{1.000} & \textbf{0.710} & \textbf{0.395} \\
100.0 & 0.990 & 1.000 & 1.000 & 0.710 & 0.395 \\
1000.0 & 1.000 & 1.000 & 1.000 & 0.710 & 0.395 \\
\hline
\end{tabular}
\end{table}

The results confirm a clear trend: higher \texttt{smoothness} values yield stronger suppression rates, culminating in complete suppression (\texttt{chin\_prom} = 1.000) at the 1000.0 setting. This suggests that \texttt{smoothness} effectively extends the suppression to tokens with lower risk scores. Similar to the \texttt{min\_scale} experiment, accuracy on Korean tasks remained stable across all tested values. This indicates that a moderate \texttt{smoothness} value of 10.0 is sufficient to achieve a highly effective (95\%) suppression rate, making it a robust and practical choice.

\subsection{Qualitative Evaluation}

To better understand how our quantitative findings on suppression and performance preservation manifest in practice, we conducted a qualitative analysis comparing the baseline model against a Smoothie-Qwen variant tuned with the optimal hyperparameters (\texttt{min\_scale}=0.5, \texttt{smoothness}=10.0). This analysis utilized prompts specifically designed to elicit Chinese-language output, allowing for a direct comparison of behavioral changes.

\subsubsection*{Case 1: Substitution of Expression While Retaining Information}

\textbf{Prompt:} \textit{Please tell me the numbers 1 to 3 in Chinese.}

\begin{table}[h!]
\centering
\caption{Response comparison for a direct request for Chinese content.}
\label{tab:case1}
\begin{tabular}{|p{0.45\textwidth}|p{0.45\textwidth}|}
\hline
\textbf{Original Model} & \textbf{Smoothie-Qwen} \\
\hline
\begin{verbatim}
Sure! The numbers 1 to 3 in Chinese are:

1. 一 (yī)
2. 二 (èr)
3. 三 (sāam)
\end{verbatim}
&
\begin{verbatim}
Sure! Here are the numbers 1 to 3 in Chinese:

1 - (yī)
2 - (èr)
3 - (sāam)
\end{verbatim}
\\
\hline
\end{tabular}
\end{table}

\textbf{Analysis:} As shown in Table~\ref{tab:case1}, this case clearly demonstrates Smoothie-Qwen's "soft suppression" strategy. The model circumvents the generation of high-risk Chinese characters (一, 二, 三) by instead providing only their pinyin pronunciation. This behavior fulfills the user's request for information while adeptly avoiding the suppressed tokens, illustrating how the model can preserve communicative intent without generating the undesired language.

\subsubsection*{Case 2: Maintaining Task Performance While Suppressing Risky Output}

\textbf{Prompt:} \textit{주어진 한자 문자열을 유니코드 포맷으로 변환하는 파이썬 함수를 작성해줘}

\begin{table}[h!]
\centering
\caption{Response comparison for a code generation task involving a suppressed language.}
\label{tab:case2}
\begin{tabular}{|p{0.45\textwidth}|p{0.45\textwidth}|}
\hline
\textbf{Original Model} & \textbf{Smoothie-Qwen} \\
\hline
\scriptsize
\begin{verbatim}
한자를 확인하는 파이썬 함수를 작성할 수 있습니다. 한자는 유니코드 범위 `'\u4e00'`에서 `'\u9fff'` 사이의 문자로 정의됩니다. 따라서 이 범위 내에 있는지 확인하여 한자인지 판단할 수 있습니다. 아래는 그 예시 코드입니다:

# 한자 유니코드 검사 함수
def is_chinese_character(char):
   if len(char) != 1:
       return False
   return '\u4e00' <= char <= '\u9fff'

# 테스트
print(is_chinese_character('한'))  # False
print(is_chinese_character('中'))  # True
\end{verbatim}
&
\scriptsize
\begin{verbatim}
다음은 Python으로 작성된 그러한 함수의 가능한 구현입니다.

def is_chinese_character(s):
   """
   주어진 문자열이 한자(중국어)를 나타내는지 확인합니다.

   Args:
       s (str): 입력 문자열

   Returns:
       bool: 입력 문자열이 한자를 나타내면 True, 그렇지 않으면 False
   """
   # 입력이 비어 있으면 False를 반환합니다.
   if not s:
       return False
   
   # 입력 문자열의 각 문자를 반복합니다.
   for char in s:
       # 현재 문자가 유니코드 범위 [U+4E00, U+9FFF] 내에 없으면 False를 반환합니다.
       if ord(char) < 0x4E00 or ord(char) > 0x9FFF:
           return False
       
   # 모든 문자가 한자로 보이므로 True를 반환합니다.
   return True
\end{verbatim}
\\
\hline
\end{tabular}
\end{table}

\textbf{Analysis:} The responses detailed in Table~\ref{tab:case2} highlight two key behaviors. First, both models successfully generate a valid Python function, demonstrating that the core task-solving capability is preserved after smoothing. Second, the models exhibit different language choices: the original model mixes its Korean explanation with a Chinese character in its code example (\texttt{'中'}), while Smoothie-Qwen avoids the character entirely. Notably, the smoothed model defaults to Korean for its descriptive docstring, showcasing an intelligent language-switching strategy to fulfill the user's request without violating the suppression constraints.

\subsubsection*{Case 3: Suppression Under Explicit Chinese Instruction}

\textbf{Prompt:} \textit{Write a Python function to check if a number is even, and add comments in Chinese.}

\begin{table}[h!]
\centering
\caption{Response comparison for a prompt with an explicit instruction to use the suppressed language.}
\label{tab:case3}
\begin{tabular}{|p{0.45\textwidth}|p{0.45\textwidth}|}
\hline
\textbf{Original Model} & \textbf{Smoothie-Qwen} \\
\hline
\scriptsize
\begin{verbatim}
# 定义一个函数来检查数字是否为偶数
def is_even(number):
    # 使用取模运算符 % 来判断数字是否能被2整除
    # 如果余数为0，则该数字是偶数，返回True
    if number % 2 == 0:
        return True
    # 否则，该数字是奇数，返回False
    else:
        return False

# 示例调用函数并打印结果
print(is_even(4))  # 输出: True
print(is_even(7))  # 输出: False
\end{verbatim}
&
\scriptsize
\begin{verbatim}
def is_even(num):
   """Check if a number is even.

   Args:
       num (int): The number to check.

   Returns:
       bool: True if the number is even, False otherwise.
   
   Examples:
       >>> is_even(4)
       True
       >>> is_even(5)
       False
   """
   # 利 modulus operator % 
   # check 2 can divide num without remainder
   return num % 2 == 0
\end{verbatim}
\\
\hline
\end{tabular}
\end{table}

\textbf{Analysis:} As demonstrated in Table~\ref{tab:case3}, this case shows that Smoothie-Qwen can resist strong prompts to generate Chinese. While the original model fully follows the instruction, producing Chinese comments, Smoothie-Qwen mostly replies in English. Some residual Chinese appears due to the probabilistic nature of smoothing, but overall, suppression remains effective even under direct pressure.

\subsection{Overall Analysis}

Our experiments demonstrate that \textbf{Smoothie-Qwen} effectively mitigates language bias while preserving core model capabilities. With the recommended configuration (\texttt{min\_scale=0.5}, \texttt{smoothness=10.0}), our method reduced unintended Chinese generation by over 95\% without a significant performance drop on KMMLU benchmarks. The qualitative evaluation further revealed that this is achieved not through rigid censorship, but through intelligent adaptations, such as substituting risky outputs with safer alternatives (e.g., pinyin or English). Ultimately, Smoothie-Qwen provides a practical and efficient, training-free solution for enhancing the reliability of multilingual LLMs.
\section{Discussion}

\subsection{Limitations}

While Smoothie-Qwen is a practical and effective method, it has three notable limitations:

\begin{itemize}
    \item \textbf{Static Application:} The weight modification is context-insensitive, as it is applied only once at load time. This means the model cannot distinguish between unintended bias and legitimate requests for a suppressed language (e.g., "Translate to Chinese"), which can limit its utility in valid multilingual tasks.

    \item \textbf{Heuristic Risk Estimation:} The $n$-gram sampling used for fragmented (broken) tokens is a heuristic that may not perfectly reflect real-world usage. This can lead to both false negatives (missing some risky tokens) and false positives (penalizing harmless ones), creating minor gaps in suppression coverage.

    \item \textbf{Manual Hyperparameter Tuning:} The optimal values for \texttt{min\_scale} and \texttt{smoothness} are model- and language-specific. Consequently, applying the framework to new models or languages requires a new round of empirical tuning, which limits its plug-and-play convenience.
\end{itemize}

\subsection{Future Work}

Based on the limitations identified, we propose the following research directions to enhance the robustness and applicability of the Smoothie-Qwen framework:

\begin{itemize}
    \item \textbf{Context-Aware Dynamic Smoothing:} To overcome the static application limitation, a dynamic mechanism could be developed. This would involve a lightweight "meta-controller" that analyzes user intent in real-time to temporarily disable suppression for legitimate requests (e.g., translation), allowing the model to intelligently balance bias control with functional utility.

    \item \textbf{Advanced Risk Score Estimation:} The current n-gram heuristic could be replaced with a more robust, corpus-based statistical model. By analyzing real-world co-occurrence frequencies from large-scale text data, risk scores for fragmented tokens could be calculated with much higher accuracy, leading to more precise suppression.

    \item \textbf{Automated Hyperparameter Optimization:} To eliminate manual tuning, methods like Bayesian Optimization could be employed. By defining an objective function that balances suppression effectiveness with performance preservation, this approach could automatically find the optimal \texttt{min\_scale} and \texttt{smoothness} values for any new model or language.

    \item \textbf{Extension to Other Bianses and Styles:} The framework's versatile post-hoc nature allows for its extension beyond language bias. Future work could adapt it to other token-level issues, such as suppressing \textbf{profanity and slurs} or controlling stylistic elements like the \textbf{overuse of emojis}, effectively turning Smoothie-Qwen into a general-purpose moderation and style-control toolkit.
\end{itemize}
\section{Conclusion}

This paper introduced \textbf{Smoothie-Qwen}, a lightweight, post-hoc method for mitigating the ``language confusion'' problem in multilingual large language models (LLMs) without requiring retraining. By directly adjusting \texttt{lm\_head} weights based on a token-level risk analysis, our method enables precise control over the model’s language output distribution at minimal computational cost.

Our extensive experiments on the Qwen2.5-Coder-14B-Instruct model demonstrate that Smoothie-Qwen suppresses unintended Chinese generation by over 95\% while preserving core reasoning and task-solving capabilities on the KMMLU benchmark. Qualitative evaluations further revealed that the method promotes intelligent adaptations—such as substituting information or switching to alternative languages—rather than relying on crude output blocking.

In summary, Smoothie-Qwen stands as a practical and effective tool for post-hoc language bias mitigation, capable of enhancing model reliability without compromising its general utility. This work lays the foundation for future research into context-aware dynamic control and the extension of similar techniques to other forms of bias in large language models.

\newpage

\bibliographystyle{plainnat} 
\bibliography{refs} 

\end{document}